\documentclass[letterpaper, 10 pt, conference]{ieeeconf}  % Comment this line out if you need a4paper
\usepackage{array}
\usepackage{times}  % DO NOT CHANGE THIS
\usepackage{helvet}  % DO NOT CHANGE THIS
\usepackage{courier}  % DO NOT CHANGE THIS
\usepackage[hyphens]{url}  % DO NOT CHANGE THIS
\usepackage{graphicx} % DO NOT CHANGE THIS
\usepackage{multirow}
\usepackage{booktabs}
\usepackage[table]{xcolor} % 添加 xcolor 宏包以支持表格中的颜色
\usepackage{xcolor}
\usepackage{cite}
\usepackage[font=footnotesize]{caption} % DO NOT CHANGE THIS AND DO NOT ADD ANY OPTIONS TO IT
\usepackage{amsthm}
\usepackage{algorithm}
\usepackage{algorithmic}
\usepackage{adjustbox}
\usepackage{newfloat}
\usepackage{listings}
\usepackage{amsfonts}
\usepackage{makecell}
\usepackage{float}
\usepackage{amsmath}
\usepackage{amssymb}
\usepackage{mathtools}
\usepackage{bm}
\usepackage[capitalize,noabbrev]{cleveref}

\theoremstyle{plain}

\theoremstyle{definition}

\theoremstyle{remark}

\newcommand{\method}{LLaVA-VLA}

\IEEEoverridecommandlockouts                              % This command is only needed if 
\title{\LARGE \bf
Rethinking the Practicality of Vision-language-action Model: \\
A Comprehensive Benchmark and An Improved Baseline
}

\author{
Wenxuan Song$^{*1}$, Jiayi Chen$^{*1}$, Xiaoquan Sun$^{*1,2}$, Huashuo Lei$^{1}$, Yikai Qin$^{1}$,\\ Wei Zhao$^{3}$, Pengxiang Ding$^{3,4}$, Han Zhao$^{3,4}$, Tongxin Wang$^{1}$, Pengxu Hou$^{1}$,\\ Zhide Zhong$^{1}$, Haodong Yan$^{1}$, Donglin Wang$^{3}$, Jun Ma$^{1}$, Haoang Li$^{1}$ \\
\\
\textbf{\textit{Codes:}} \url{https://github.com/OpenHelix-Team/LLaVA-VLA}
}
\begin{document}
\twocolumn[{%
\renewcommand\twocolumn[1][]{#1}%
\maketitle
\begin{center}
    \centering
    \captionsetup{type=figure}
    \includegraphics[width=0.99\linewidth]{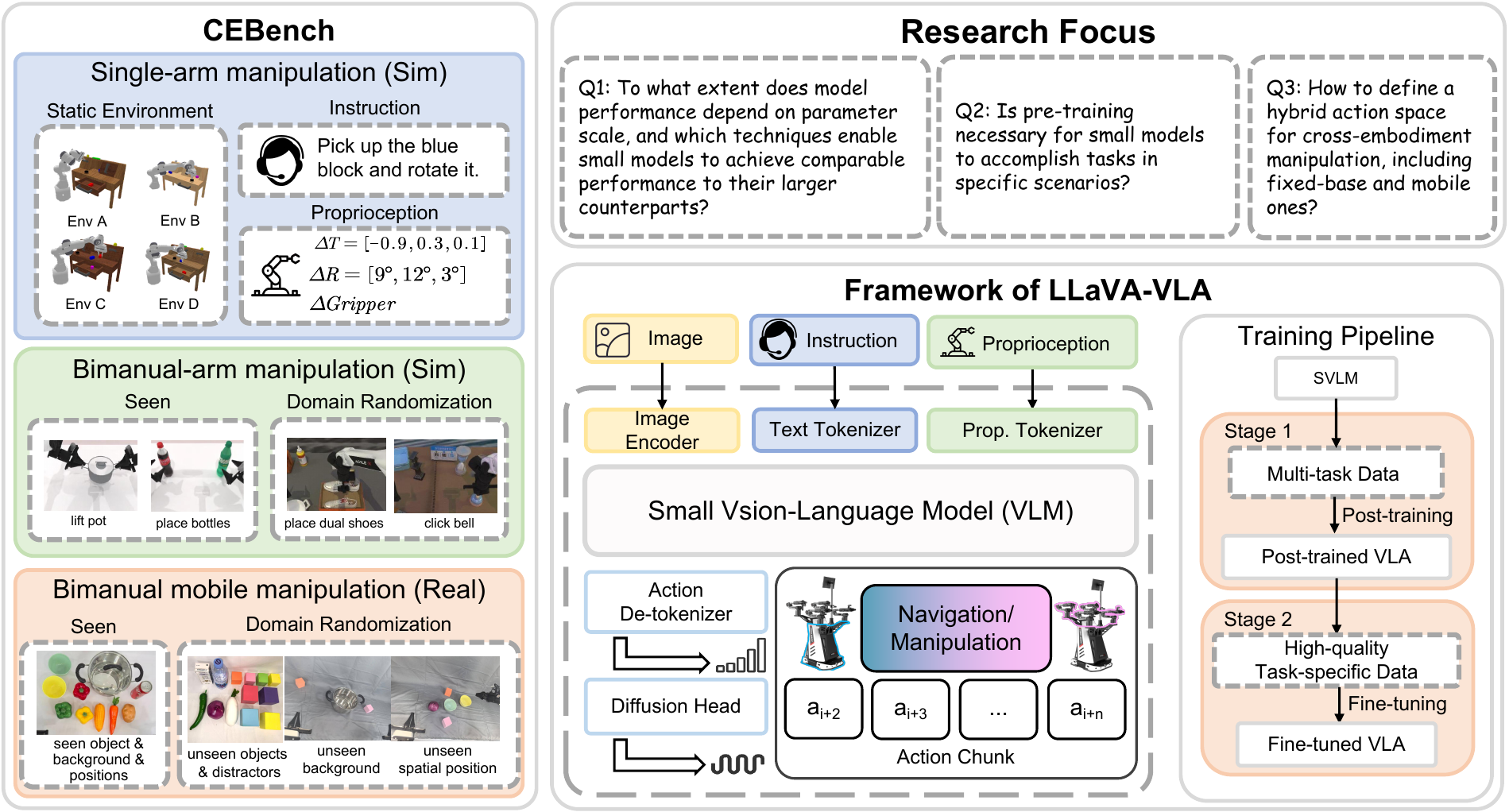}
    \captionsetup{font=footnotesize}
    \captionof{figure}{
    Overview of this work. We conduct a comprehensive study on the practicality of vision-language-action models.
    We first construct a cross-embodiment benchmark, CEBench, across simulation and the real world, and offer diverse evaluation settings. 
    Then we explore three critical aspects (Q1-Q3) and offer several key findings.
    Based on the above findings, we introduce our \method, a lightweight yet effective baseline capable of mobile manipulation.
    }
    \label{fig:teaser}
    \vspace{1.5em}
\end{center}
}]
% \maketitle
% \thispagestyle{empty}
% \pagestyle{empty}%\footnotemark

\begingroup
\makeatletter
\renewcommand{\@makefntext}[1]{\noindent #1} % 关键：取消ieeeconf脚注预留的标记缩进
\makeatother
\renewcommand{\thefootnote}{}

\footnotetext{%
*Wenxuan Song, Jiayi Chen and Xiaoquan Sun contributed equally to this work.\\
\textsuperscript{1}The Hong Kong University of Science and Technology (Guangzhou), Guangzhou, China.\\
\textsuperscript{2}Huazhong University of Science and Technology, Wuhan, China.\\
\textsuperscript{3}Westlake University, Hangzhou, China.\\
\textsuperscript{4}Zhejiang University, Hangzhou, China.\\
}
\endgroup
\setcounter{footnote}{0}

%%%%%%%%%%%%%%%%%%%%%%%%%%%%%%%%%%%%%%%%%%%%%%%%%%%%%%%%%%%%%%%%%%%%%%%%%%%%%%%%
\begin{abstract}
% The emergence of Vision-Language-Action (VLA) models has revolutionized visuomotor control by enabling robots to interpret multimodal sensory inputs and follow complex instructions through end-to-end learning. 
Vision-Language-Action (VLA) models have emerged as a generalist robotic agent.
However, existing VLAs are hindered by excessive parameter scales, prohibitive pre-training requirements, and limited applicability to diverse embodiments. 
To improve the practicality of VLAs, we propose a comprehensive benchmark and an improved baseline.
% To provide practicality evaluation and systematically explore VLAs, we propose a comprehensive benchmark spanning diverse embodiments in both simulation and the real world  and an improved baseline.
First, we propose CEBench, a new benchmark spanning diverse embodiments in both simulation and the real world with consideration of domain randomization.
We collect 14.4k simulated trajectories and 1.6k real-world expert-curated trajectories to support training on CEBench.
Second, using CEBench as our testbed, we study three critical aspects of VLAs' practicality and offer several key findings.
Informed by these findings, we introduce \method, a lightweight yet powerful VLA designed for practical deployment on consumer-grade GPUs. 
Architecturally, it integrates a compact VLM backbone with multi-view perception, proprioceptive tokenization, and action chunking.
To eliminate reliance on costly pre-training, \method~adopts a two-stage training paradigm including post-training and fine-tuning. 
Furthermore, \method~extends the action space to unify navigation and manipulation. 
% To support systematic evaluation, we present  
Experiments across embodiments demonstrate the capabilities of generalization and versatility of \method~, while real-world mobile manipulation experiments establish it as the first end-to-end VLA model for mobile manipulation. 
% This work takes a significant step toward democratizing VLA research by making such models lightweight, robust, and accessible for mobile and embodied robotic systems. 
We will open-source all datasets, codes, and checkpoints upon acceptance to foster reproducibility and future research.
\end{abstract}

\renewcommand{\thefootnote}{}
\noindent

% \textsuperscript{5}Harbin Institute of Technology, Harbin, China.}
% \vspace{-0.5cm}
%%%%%%%%%%%%%%%%%%%%%%%%%%%%%%%%%%%%%%%%%%%%%%%%%%%%%%%%%%%%%%%%%%%%%%%%%%%%%%%%
\section{INTRODUCTION}

%第一段
The emergence of Vision-Language-Action     (VLA)~\cite{rt1,gr1,black2024pi_0, PD-VLA, CEED-VLA,quarvla,song2025rationalvla,reconvla,udvla,upvla,spatialforcing} models has revolutionized the field of robotics, offering powerful capabilities in visuomotor control and the comprehension of complex instructions through end-to-end learning processes. These models have demonstrated their potential in various robotic tasks, enabling systems to interpret multimodal sensory data and execute actions based on language instructions. 
However, despite their success, current VLAs face significant hurdles that impede their widespread deployment and practical use in real-world scenarios:
%通往真实世界的vla模型受到以下几个关键问题的制约: 1. 参数量太大，难以向移动端硬件上部署 2. 预训练太麻烦 3. 现有的vla模型无法满足需要移动位置并进行操纵的任务
% One of the primary challenges of VLA model is their sheer scale, 
1) \textbf{Billions of parameters} make them difficult to deploy in resource-constrained environments, such as mobile platforms and consumer-grade devices.
2) \textbf{Extensive pre-training} using large-scale robotic datasets leads to prohibitive training costs and the need for vast computational resources.
3) \textbf{Fixed-base manipulation} limits their applicability to cross-embodiment deployment, \textit{i.e.}, mobile manipulation tasks.

%第二段
%一些文章已经在一定程度上探索了模型轻量化的问题，然而，他们1. 缺乏在多个构型上的系统性探究，2. 并且仍然无法胜任移动操纵的任务。
% Recent works have explored compact VLA architectures~\cite{tinyvla,smolvla,hung2025nora}. 
Existing work has partly investigated some of the aforementioned issues.
TinyVLA~\cite{tinyvla} introduced a 1B-level model trained from scratch.
It employs Low-Rank Adaption~\cite{hu2021lora} and diffusion head for efficient training and inference.
% but the inclusion of a diffusion head increased architectural complexity and impaired autoregressive reasoning. 
% MiniVLA~\cite{belkhale2024minivla} is a variant of OpenVLA~\cite{kim2024openvla} with a smaller backbone and uses RVQ-VAE~\cite{vqvae} to quantize actions, achieving fast and precise inference. 
MiniVLA~\cite{belkhale2024minivla} is a variant of OpenVLA~\cite{kim2024openvla} with a smaller backbone and uses a  VQ-VAE tokenizer to quantize actions, achieving fast and precise inference. 
NORA~\cite{hung2025nora} utilizes FAST~\cite{fast} to quantize action tokens and diffusion expert.
SmolVLA~\cite{smolvla} realizes efficient training through skipping layers, pruning visual tokens, and initializing a small VLM backbone.
% These attempts highlight an ongoing trade-off between scale reduction, architectural simplicity, and performance.
%尽管取得了进展，但这些方法没有进行系统性探究以考虑每部分设计的作用，同时没有在不同构型的机器人上验证其有效性，并且仍然无法胜任移动操纵的任务。
Despite these advances, these approaches have not systematically examined the practicality of their lightweight and pretraining-free designs, and they remain incapable of performing mobile manipulation tasks.

%第三段
%为了解决这些问题，我们首先构建了跨构型的benchmark，它包含仿真和真实中的多个场景和任务，并考虑了真实的域随机化。基于这个bench，我们探索了四个问题，并给出了七个关键发现。
To improve the practicality of VLAs, we propose a comprehensive benchmark and an improved baseline.
First, we introduce CEBench, a practical robotic benchmark suite that spans diverse embodiments (single arm~\cite{calvin}, bimanual~\cite{robotwin}, and mobile bimanual) in both simulation and the real world, with explicit consideration of domain randomization. 
In CEBench, we collect 14.4k simulated demonstrations across 36 tasks and 1.6k high-quality real-world demonstrations across 8 tasks.

Second, using CEBench as our testboard, we study three critical aspects for VLAs: the lightweight designs, the training curriculum, and the unified action space, and empirically offer several valuable findings and insights into their choices.
%第四段
%基于这些发现，我们提出了LightVLA，他保持了轻量化的模型结构，强大的性能，以及突破性的移动操纵能力。1.模型结构集成了xxxx技术，2.通过post-training取代了pre-training，3. 并将动作空间扩展至移动双臂操纵。实验表明xxxxxxx
Informed by these findings, we propose~\method, a \textbf{lightweight} VLA with strong performance, which is capable of \textbf{mobile manipulation}.
\Cref{fig:teaser} shows that
~\method~does \textbf{not need pre-training} and can be trained and deployed on consumer-grade GPUs.
~\method~initializes a compact VLM and integrates multi-view input, proprioception tokenization, and action chunking to balance efficiency and performance while maintaining minimalist model design. 
To eliminate the reliance on pre-training, we further adopt a two-stage training paradigm, combining post-training on multi-task data with fine-tuning on task-specific data. 
Finally, we design a unified action space including a specific action space for navigation and a manipulation action space combined through several special tokens.
% Its efficiency stems from three key design principles: incorporating proprioceptive signals to ground the model in the robot’s embodied state; leveraging multi-view images to capture richer 3D spatial understanding; and introducing action chunking to stabilize long-horizon planning and reduce error accumulation.

Extensive evaluation of our \method~on CEBench shows that it matches or surpasses models over 10× larger, especially under domain-randomized settings, which demonstrates its strong visual generalization capabilities.
% 1. 超过更大模型的性能 2. 良好的视觉泛化性
%多个embodiments上的实验证明了我们模型的通用性和混合动作空间设计的有效性
%进一步，真机实验证明了lightvla是首个可用于移动manipulation的端到端vla。
% Results on bimanual in RoboTwin show that \method~achieves higher success rates on tasks with domain randomization, indicating stronger capabilities of generalization compared with pre-trained, larger robotic foundation model RDT~\cite{rdt}. 
% Results on the mobile manipulator in the real world show that our 
% coordinate navigation and manipulation and
\method~successfully manages tasks like \textit{move to the operation table from outside and place the bottle},
which indicates that \method~is the first end-to-end VLA model capable of handling mobile manipulation tasks.
Cross-embodiment experiments demonstrate the versatility of our model and the effectiveness of the proposed hybrid action space design. 
% Furthermore, real-world robotic evaluations verify that LightVLA represents the first end-to-end VLA model capable of handling mobile manipulation tasks.
This work represents a promising step toward democratizing VLAs by making them lightweight, generalized, and practical for deployment in mobile embodied systems. 
% We release all datasets, codes, and checkpoints to provide a reference for future research in open-source VLAs.
% %看情况可以不留了

To summarize, our key contributions are:
\begin{itemize}
    \item We construct a benchmark to evaluate the practicality of VLAs, which spins diverse embodiments in both simulation and the real world, and considers domain randomization.
    \item We conduct a comprehensive study on the practicality of VLAs and offer several insightful findings.
    \item We propose an improved baseline, \method, which is lightweight, pretraining-free, and capable of mobile manipulation.
    \item We will release all datasets, codes, and checkpoints upon acceptance to provide a reference for future research in open-source VLAs.
%     \item We validate the generality and effectiveness of our method through cross-embodiment experiments in both simulation and real-world settings.
%     \item We open-source the full \method~ framework, including model checkpoints, training strategy, and evaluation protocols, to facilitate reproducibility and promote further research in scalable vision-language-action models for robotics.
\end{itemize}

\section{RELATED WORKS}
\subsection{Vision-Language Models}
% llava opemfaminfo qwen emu3 clip siglip vae show-o prismatic
% 基于llama qwen 等大预言模型的发展，llava opemfaminfo prismatic结合clip和siglip在ebeding纬度融合视觉信息到文本空间，经过大规模多阶段的视觉语言对齐训练，使得vlm模型具有很好的视觉推理能力。同时 show-o emu3 mmada 结合MAGVIT-v2 [24] SBER-MoVQGAN在编码成离散token，从而在词元维度上统一了图像token，同时拥有不错的视觉推理和生成能力。最近 llava-next设计三阶段的训练策略进一步提升了vlm的性能 gpt, internvl, deepseek, ross，minicpm
Vision-Language Models (VLMs) \cite{minigpt-4,gemini,achiam2023gpt} have rapidly advanced and extended the success of Large Language Models (LLMs)~\cite{llama,bai2023qwen} into multimodal domains. By incorporating visual modalities, these models achieve impressive performance on tasks requiring visual reasoning and understanding.
At the embedding level, a representative line of work leverages vision encoders such as CLIP~\cite{clip} or SigLIP~\cite{siglip} to embed visual information into the language space. Through large-scale multistage alignment training~\cite{prismatic-vlms,llava}, these models significantly enhance cross-modal alignment and exhibit strong visual reasoning abilities.
% In parallel, another line of research~\cite{show-o,emu3} unifies visual and textual modalities at the token level, enhancing reasoning while enabling strong image generative abilities.
These developments in language and vision further inspire extensions to other modalities, including action, touch, and sound.  
% Furthermore,
% \cite{wangreconstructive}引入重建模块在训练时监督图像重建ross，帮助模型更好的理解图像

% VLMs bridge vision and language, extending the reasoning capabilities of large language models (LLMs) to multimodal input~\cite{minigpt-4,vit,gemini,minigpt-v2}. Conventional approaches typically integrate a pre-trained vision encoder~\cite{prevision1,prevision2} with a pre-trained LLM~\cite{related1,prellm1,prellm2}, followed by large-scale image-text pre-training and instruction tuning. However, most existing VLMs contain 7B–70B parameters, resulting in high inference costs and limited accessibility. Therefore, recent efforts focus on lightweight VLMs (3B parameters), unified architectures that represent both images and text as discrete tokens, and parameter-efficient tuning strategies to improve training and inference efficiency. 
% Beyond images and text, VLA research is expanding to action, further broadening the scope of robot control.

\subsection{VLA Models}
% ReconVLA flowvla  MoRE Germ openhelix rationalvla pdvla ceedvla Quar-vla
% rt1 rt2 openvla gr1 grootn1 ocoto 
% Early works such as RT-1~\cite{rt1} and Octo~\cite{octo_2023} trained transformer models from scratch on large-scale robot trajectories to learn end-to-end action policies. RT-2~\cite{rt2} extended this paradigm by incorporating web-scale vision-language data, enhancing generalization beyond robot-specific datasets. RoboFlamingo~\cite{li2024roboflamingo} adapted pre-trained VLMs to robotic manipulation through lightweight imitation learning, achieving state-of-the-art performance with low deployment cost. OpenVLA~\cite{kim2024openvla} further introduced the first open-source 7B-parameter VLA model, trained entirely on public data and capable of generating discrete action tokens.
% $\pi0$~\cite{black2024pi_0} combined flow-matching diffusion with action chunking, replacing discrete tokenization and enabling efficient high-frequency action generation.  
Early works such as~\cite{rt1,gr1,black2024pi_0} trained transformers from scratch on web-scale vision-language data and large-scale robot trajectories to improve performance and generalization. RoboFlamingo~\cite{li2024roboflamingo} adapted pre-trained VLMs to manipulation via lightweight imitation learning.
OpenVLA~\cite{kim2024openvla} released the first open-source 7B VLA trained on public data. OpenHelix~\cite{openhelix} proposed a dual-system dual-system architecture. 
More recently, GR00T N1~\cite{bjorck2025gr00t} advances general-purpose humanoid control with a diffusion-based action generator in a dual-system architecture.
% $\pi0.5$\cite{intelligence2025pi_} enhances generalization and data integration through a hierarchical reasoning architecture and hybrid action representation. 
PD-VLA~\cite{PD-VLA} and CEED-VLA~\cite{CEED-VLA} explored the acceleration of inference.
ReconVLA~\cite{reconvla} and FlowVLA~\cite{zhong2025flowvla} leveraged extra low-level visual perception. VLA-Adapter~\cite{vla-adapter} proposes a lightweight, parameter-efficient bridge from pre-trained VLMs to action prediction. VLM4VLA~\cite{zhang2026vlm4vla} benchmarks how VLM choice affects VLA performance with minimal adaptation.
However, these models remain computationally demanding and pose challenges for deployment in resource-constrained settings and mobile manipulation tasks.

\section{CEBench}
%xiaoquan8.31
%先前的方法缺乏跨实体、sim-to-real的系统化实验。为了弥补这一差距，我们提出了CEBench，一个可靠的基准测试套件，用于全面研究端侧模型的部署效果。
Prior benchmarks exhibit a significant gap in practical deployment, lacking comprehensive and unbiased evaluation across embodiments, potential domain randomization, and mobile manipulation needs. 
To bridge this gap, we propose Cross-Embodiment Benchmark (\textbf{CEBench}), a reliable benchmarking suite designed for systematic evaluation of practicality in VLAs.
Notably, we treat CALVIN~\cite{calvin} as a subset of our dataset and mainly introduce our dataset in RoboTwin~\cite{robotwin} and the real world.

\subsection{System Setup}
%介绍robotwin, calvin和真机。
We evaluate our~\method~in both simulated and real-world environments to comprehensively assess task performance and generalization. 

\noindent
\textbf{Single-arm manipulation.} 
The CALVIN benchmark\cite{calvin} includes a Franka Panda single robotic arm and a table environment to study long-horizon language-conditioned manipulation and visual generalization.
% consists of 34 tasks across four environments (A, B, C, and D) involving a Franka Panda single robotic arm.
% CALVIN is designed to study long-horizon language-conditioned manipulation tasks by setting a task sequence including 5 sub-tasks. 
% Unlike existing vision-and-language datasets, CALVIN introduces higher complexity in terms of sequence length, action space, and linguistic variability.
% We select the challenging ABC→D setup to test the generalization on unseen backgrounds.

\noindent
\textbf{Bimanual manipulation.} 
The RoboTwin benchmark\cite{robotwin}, built on the Sapien~\cite{xiang2020sapien} simulator, designed to evaluate positional generalization and visual robustness.
It fosters an expert data synthesis pipeline that leverages VLMs and simulation-in-the-loop refinement to automatically generate task-level execution code.

\begin{figure}[t]  
    \centering
    \includegraphics[width=0.99\linewidth]{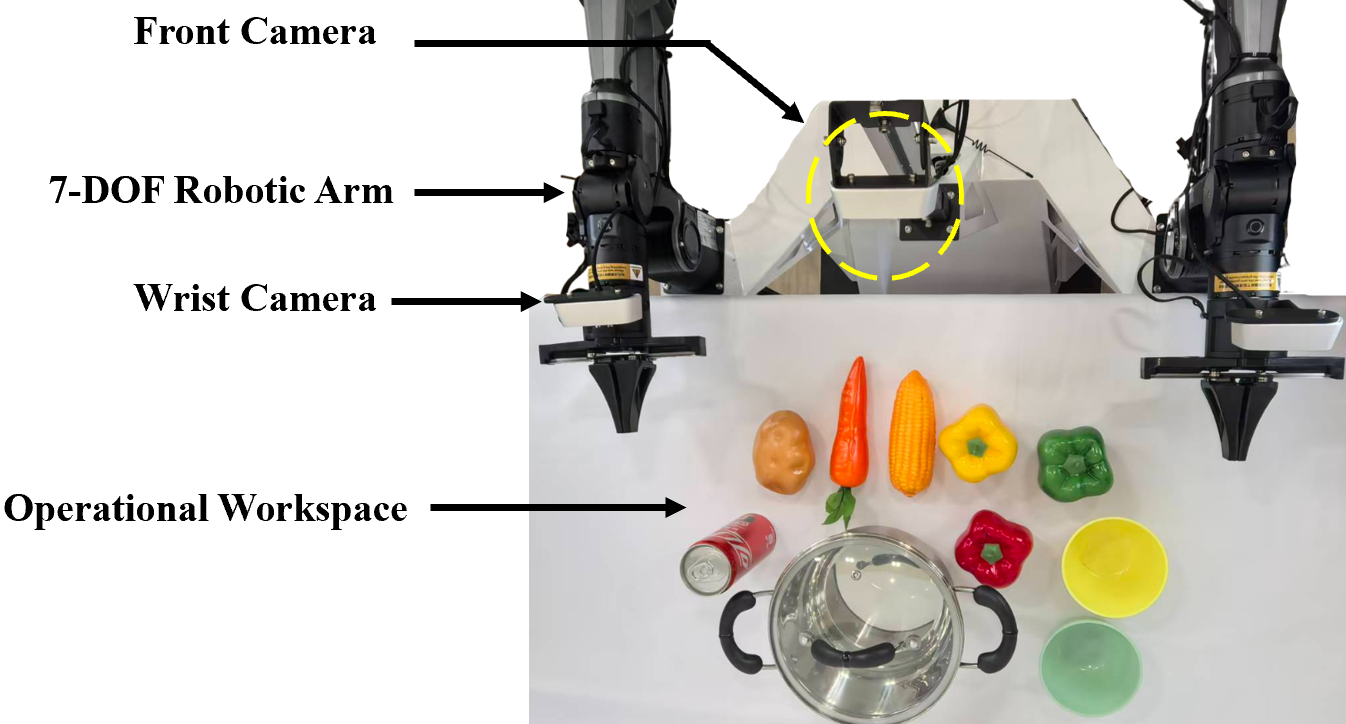} 
    \caption{Real-world setup of the Cobot-Magic system for mobile bimanual manipulation (top view).}
    \label{fig:example} 
\end{figure}

\noindent
\textbf{Bimanual mobile manipulation.} 
\Cref{fig:example} shows the real-world experimental setup using the Cobot-Magic dual-arm mobile robot, equipped with four Piper robotic arms (two master arms and two puppet arms), which capture RGB images at a resolution of 480 × 640 and 30 Hz.
% The real-world part of our benchmark comprises 10 tasks, such as \textit{stack bowls} and \textit{place bottles}, covering a variety of fine-grained manipulation scenarios.

\subsection{Datasets}
%我们在robotwin中收集了多少数据和任务，真机收集了多少数据和任务(按10个写)
For CALVIN, we use the official datasets.
On the RoboTwin platform~\cite{robotwin}, we constructed a large-scale simulation dataset containing 14.4k trajectories and 36 tasks (400 trajectories per task) in an automatic manner. 
All trajectories were collected under a simplified scenario with a clutter-free tabletop and stable background and lighting. 
In the real world, we designed 8 tasks and collected 200 trajectories per task. 
The tasks include:
\begin{itemize}
    \item \textbf{Stack bowls}: Pick the bowl and put it on the other.
    \item \textbf{Restore bottle}: Return a toppled bottle to its upright position.  
    \item \textbf{Click bell}: Trigger a desk bell by pressing its small button.  
    \item \textbf{Place vegetable}: Grasp a vegetable and correctly place it into the target container.  
    \item \textbf{Pack bottles}: Insert two bottles neatly into a designated box.  
    \item \textbf{Lift pot}: Grasp a pot with two arms and lift it off the table surface.  
\end{itemize}
These tasks cover different difficulties, from basic pick-and-place operations to more complex bimanual interactions.
% Among them, \textit{click bell}, \textit{lift pot}, and \textit{move can pot} are used for sim-to-real experiments on the RoboTwin platform. 
We also collect 2 mobile manipulation tasks:
\begin{itemize}
    \item \textbf{Move and fetch bottles}: move to the table and fetch the bottle on it.
    \item \textbf{Move and open the drawer}: move to the drawer and open it.
\end{itemize}
% To evaluate model generalization, we conducted systematic evaluation on eight tasks within the RoboTwin environment: 
These data provide a solid foundation for systematic training and evaluation.

\subsection{Evaluation and Metrics}
%通过研究仿真和真机相同任务下的成功率，我们验证了由数字孪生驱动的仿真器对于真机性能的可靠反应。
%为了衡量模型在端侧部署的效率，我们额外引入了FLOPs(计算量)和最低硬件要求。
For CALVIN, we follow the official evaluation setting and report the success rates of each sub-task as well as the average success length across 5 tasks.
For RoboTwin, we select 8 representative tasks to ensure fair and efficient comparison: \textit{click bell}, \textit{click alarmclock}, \textit{lift pot}, \textit{move can pot}, \textit{open laptop}, \textit{pick dual bottles}, \textit{place dual shoes}, \textit{rotate qrcode}.
The evaluation on RoboTwin is conducted on \textbf{seen} settings as well as unseen settings with domain randomization (\textbf{DR}), which includes clutter, random lighting, diverse textures, and variable table heights. 
For real-world experiments, we evaluate on all tasks in the datasets.
To simulate the DR setting in reality, we vary the color and texture of the tabletop and randomly place distractor objects (e.g., \textit{blocks, pens}) in the workspace, creating visual and layout variations that were unseen in training.
% For both simulated and real-world environments, we implement \textbf{Seen} settings as well as unseen settings with domain randomization (\textbf{DR}).
% For RoboTwin, we also adopt the official evaluation setup, using both \textbf{Easy} and \textbf{Hard} settings in simulation and the real world. 
% In RoboTwin, the DR setting is with clutter, random lighting, diverse textures, and variable table heights. 
% By examining the success rates of both simulated and physical robots under identical tasks, we validate that the simulator driven by digital twins reliably reflects the performance of the physical system.
% To further assess deployment efficiency on edge devices, we include FLOPs (floating-point operations) and minimum hardware requirements as key metrics.

%记得介绍clean和域随机化两种设定(已完成)

\subsection{Baselines} 
To comprehensively evaluate the performance of policies and VLAs with fewer than 1B parameters, we conduct comparisons with the following baselines in RoboTwin and the real world in \Cref{sec:final}.
\begin{itemize}
    \item \textbf{ACT~\cite{act}}: A CVAE-based imitation learning approach is proposed, which leverages action chunking to forecast sequences of upcoming actions and incorporates temporal ensembling to achieve smooth execution.
    \item \textbf{Difussion policy~\cite{dp}}: A visuomotor policy learning framework that models action prediction through a conditional denoising diffusion process.
    \item \textbf{TinyVLA~\cite{tinyvla}}: A compact VLA model that leverages a lightweight multimodal backbone to efficiently generate robot actions from vision-language inputs, enabling fast inference and strong generalization across diverse tasks.
    \item \textbf{RDT~\cite{rdt}}: A Transformer model incorporating diffusion is proposed for bimanual robotic manipulation, which employs a unified action space and multimodal inputs to enable efficient few-shot learning across various tasks.
\end{itemize}
%对于CALVIN，我们选择其leaderboard上几种不同类型的代表性方法。
For the CALVIN benchmark, we compare with representative methods on the official leaderboard.

\section{\method~and Its Associated Findings}
% Current VLA models exhibit substantial variation in architectures, input, output, model, and training paradigms.
% both the selection and training paradigms of key components, including base vision language models, downstream policy architectures, latent representations, and asynchronous strategies for training and inference. 
To build a lightweight, pretraining-free, and cross-embodiment VLA for practical use, we systematically explore design choices of VLAs.
Specifically, we formulate three research questions to guide our explorations:

% \textbf{Q1:} Which techniques are critical for improving the performance of small models?
\textbf{Q1:} To what extent does model performance depend on parameter scale, and which techniques enable small models to achieve comparable performance to their larger counterparts?

\textbf{Q2:} Is pre-training necessary for small models to accomplish tasks in specific scenarios?

% \textbf{Q3:} Is model performance strictly proportional to parameter scale, or are small VLAs inherently much weaker than large VLAs?

% \textbf{Q4:} Is the diffusion head indispensable for effective policy generation?
% \noindent
\textbf{Q3:} How to define a unified action space for cross-embodiment manipulation, including fixed-base and mobile ones?

% Each question is conducted in a single environment in CEBench to consider practical constraints on computational resources during research. 

% Q1. 小模型必须预训练才能完成特定场景下的任务吗？
% Q2. 哪些技术是提升小模型性能的关键？
% Q3. 模型的性能与参数量成正比吗/小型VLA是否远弱于大型VLA？
% Q4. diffusion head是必需的吗？

%通过一系列实验，我们得到了以下key findings:
% 1. 几个关键技术让小模型可以媲美大模型: 多视角、本体感知、ac
% 2. 多视角简单拼接比分开输入效果好
% 3. 参数量的影响比预想的更小
% 4. 一些技术容易被高估：diffusion head不比discrete强太多、chunk bin的精度不会带来性能提升、moe没用、
% 5. 大规模预训练在单一任务场景下不是必须的

\begin{figure}[t]  
    \centering
    \includegraphics[width=0.99\linewidth]{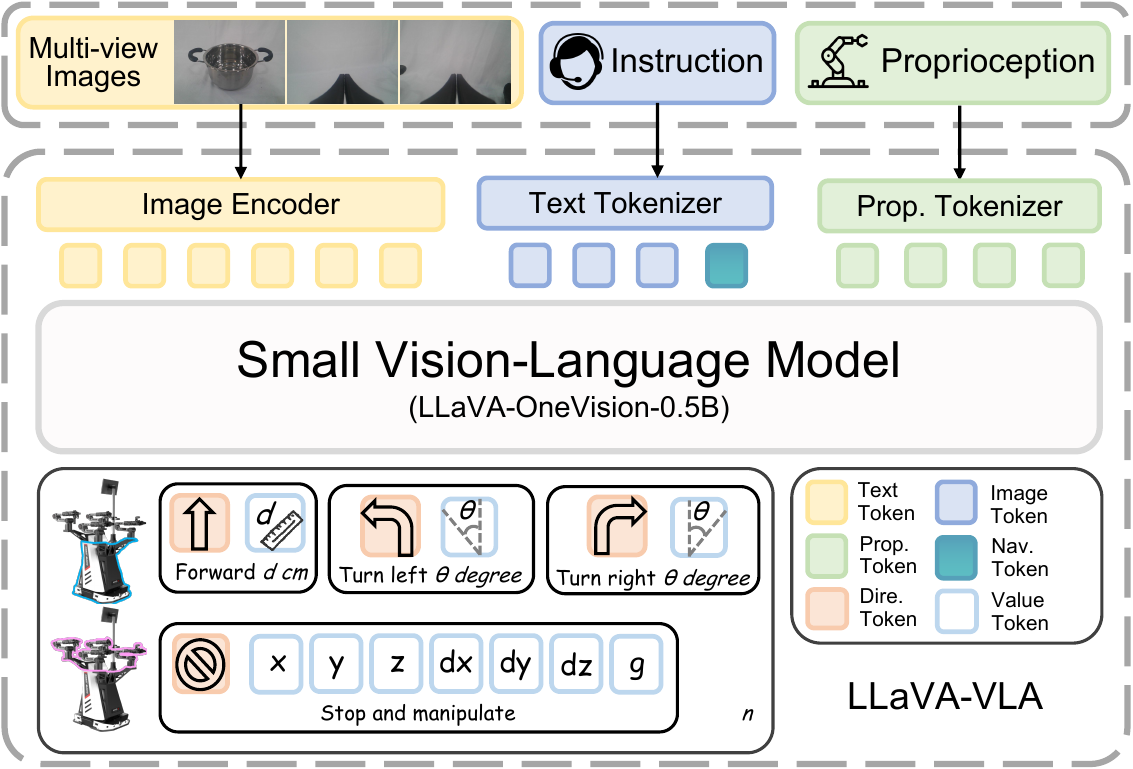} 
    \caption{Model architecture of our \method.}
    \label{fig:model} 
\end{figure}

\subsection{\method}
%我们以一个至上而下的顺序进行深入研究，即我们首先给出完整的模型，再依次介绍我们在建立模型中的关键发现和设计。从一个预训练的LLaVA-OneVision-0.5b作为backbone。并使用拼接的多视角图片作为输入、本体感知作为额外的输入，通过action tokenizer输出离散的动作块。
We conduct our study in a top-down manner.
We first present our \textbf{Light}weight \textbf{VLA} (\method). 
As shown in \Cref{fig:model}, \method~is built upon a pre-trained LLaVA-OneVision-0.5B~\cite{llavaov} backbone, taking concatenated multi-view images as input, augmented with proprioceptive signals, and producing action chunks through an action tokenizer.
%为了实现移动操纵的能力，我们构建了hybrid action space，其中包含direction token和对应的value token，这允许模型在navigation和manipulation之间自由切换。
To enable mobile manipulation, we construct a hybrid action space consisting of direction tokens and their corresponding value tokens, which allows the model to seamlessly switch between navigation and manipulation.
In the following part, we discuss the design choices made during its development step by step to provide the key findings (F1-F8).

\subsection{Study on Lightweight Designs (Q1)}
%为了实现轻量级的vla模型，我们首先探讨轻量级模型的性能是否有望媲美大模型，紧接着给出实现这个级别性能的关键设计。
Toward lightweight VLAs, we first investigate whether compact models can rival large-scale counterparts (F1), then identify the key design choices that make such performance attainable (F2-4).

\noindent
\textbf{Key Findings 1: Model performance is not strictly proportional to parameter scale. 
Small VLAs can achieve performance comparable to their large-scale counterparts.}
\Cref{tab:parameters}, \method-0.5B achieves performance on par with 7B models despite less than 10\% parameters. 
On the first sub-task, it reaches a success rate of 96.2\%, comparable to 97.4\% of its 7B counterpart.
On the last task, it achieves 50.6\%, significantly outperforming 23.5\% of the 3B RoboFlamingo and 43.5\% of the 7B OpenVLA. 
With an average action length of 3.65, it fully surpasses these larger models. 
These results suggest that performance gains are not solely determined by scale.
With the proposed techniques, small VLAs can match or surpass larger models.

\begin{table}[H]
\caption{Comparison among different versions of \method~in terms of success rates and average length on RoboTwin. Here, B denotes billions.}
\centering
\small % 或 \footnotesize 进一步缩小
\setlength{\tabcolsep}{2.3pt} % 减小列间距（默认是 6pt）

\label{tab:parameters}
\begin{tabular}{cccccccc}
\toprule
\multirow{2}{*}{Model} & \multirow{2}{*}{Param.}&
\multicolumn{5}{c}{Success Rate (\%)} & Avg. Len. \\
& & 1/5 & 2/5 & 3/5 & 4/5 & 5/5 & ABC$\rightarrow$D \\
\midrule
\method~(ours) & 0.5B & 96.2 & 84.8 & 72.6 & 60.8 & 50.6 & 3.65 \\
\method~(ours) & 7B & 97.4 & 86.2 & 73.4 & 64.6 & 53.4 & 3.75 \\
\midrule
\textcolor{gray!80}{RoboFlamingo~\cite{li2024roboflamingo}} & 
\textcolor{gray!80}{3B} & 
\textcolor{gray!80}{82.4} & 
\textcolor{gray!80}{61.9} & 
\textcolor{gray!80}{46.6} & 
\textcolor{gray!80}{33.1} & 
\textcolor{gray!80}{23.5} & 
\textcolor{gray!80}{2.47} \\
\textcolor{gray!80}{OpenVLA~\cite{kim2024openvla}} & 
\textcolor{gray!80}{7B} & 
\textcolor{gray!80}{91.3} & 
\textcolor{gray!80}{77.8} & 
\textcolor{gray!80}{62.0} & 
\textcolor{gray!80}{52.1} & 
\textcolor{gray!80}{43.5} & 
\textcolor{gray!80}{3.27} \\
\bottomrule
\end{tabular}

\end{table}

\noindent
\textbf{Key Findings 2: Multi-view images are critical as they enable stereoscopic perception of 3D space, and containing multi-view information in 1 image is an effective way.}
%这一节需要跟一个与直接拼接多图的比较实验@swx
In manipulation tasks, third-person view images provide global contextual information, while first-person view images offer precise object-to-gripper positional cues, which are crucial for achieving precise manipulation. 
While inheriting the above information, multi-view images capture disparity information, which is essential for constructing a three-dimensional understanding of the scene.
Therefore, incorporating both perspectives is essential. 

Several strategies exist for handling multi-view inputs~\cite{robovlm, oft}. 
Encoding each image separately and then concatenating its image tokens typically leads to an excessive number of image tokens and introduces considerable redundancy, resulting in suboptimal performance. 
One potential remedy is to apply token compression methods to reduce visual token count. 
However, this approach may incur information loss, which may result in slight performance degradation. 
Consequently, we adopt a simpler yet effective strategy: vertically concatenating the first- and third-person view images into a single composite image. 
This approach not only reduces the number of tokens while preserving complete multi-view visual information, but also aligns with the training paradigm of our VLM backbone, thereby avoiding potential performance degradation.
\begin{table}[H]
\caption{Comparison of different methods for integrating multi-view images on CALVIN.}
\centering
\small % 或 \footnotesize 进一步缩小
\setlength{\tabcolsep}{2.3pt} % 减小列间距（默认是 6pt）
\label{tab: proprioception}
\begin{tabular}{cccccccc}
\toprule
% VLM & Param. &
% \multicolumn{5}{c}{Success Rate (\%)} & Avg. Len. \\
% method & projector & 1/5 & 2/5 & 3/5 & 4/5 & 5/5 & ABC$\rightarrow$D \\
\multirow{2}{*}{Method} & \multicolumn{5}{c}{Success Rate (\%)} & Avg. Len. 
\\    & 1/5 & 2/5 & 3/5 & 4/5 & 5/5 & ABC$\rightarrow$D \\
\midrule
Concate Image Tokens & 65.3 & 37.1 & 23.5 & 14.7 & 9.2 & 1.50 \\
Merged Image & \textbf{94.8} & \textbf{84.5} & \textbf{71.3} & \textbf{62.5} & \textbf{53.8} & \textbf{3.68} \\ 
\bottomrule
\end{tabular}

\end{table}

\noindent
\textbf{Key Findings 3: Proprioception is critical as it improves understanding of physical states, and tokenizing proprioception works better than encoding them by linear layers.}
Proprioceptive information is critical for enabling robots to infer their current state and maintain action continuity. 
A common approach is to encode this information using an MLP. 
In our design, we translate proprioception values into a sequence of proprioception tokens via a proprioception tokenizer, which can be regarded as an inverse form of the action de-tokenizer. 
As shown in \Cref{tab: proprioception}, this integration facilitates better exploitation of the VLM’s language modeling capabilities for understanding and generating coherent actions.
\begin{table}[H]
\caption{Comparison of different methods for integrating proprioceptive information on CALVIN.}
\centering
\small % 或 \footnotesize 进一步缩小
\setlength{\tabcolsep}{2.3pt} % 减小列间距（默认是 6pt）

\label{tab: proprioception}
\begin{tabular}{cccccccc}
\toprule
% VLM & Param. &
% \multicolumn{5}{c}{Success Rate (\%)} & Avg. Len. \\
% method & projector & 1/5 & 2/5 & 3/5 & 4/5 & 5/5 & ABC$\rightarrow$D \\
\multirow{2}{*}{Method} & \multicolumn{5}{c}{Success Rate (\%)} & Avg. Len. 
\\    & 1/5 & 2/5 & 3/5 & 4/5 & 5/5 & ABC$\rightarrow$D \\
\midrule
 MLP Projector & 90.4 & 76.0 & 58.0 & 48.0 & 37.2 & 3.09 \\
Prop. Tokenizer & \textbf{94.8} & \textbf{84.5} & \textbf{71.3} & \textbf{62.5} & \textbf{53.8} & \textbf{3.68} \\ 
\bottomrule
\end{tabular}

\end{table}

\noindent
\textbf{Key Findings 4: Action chunking is critical as it strengthens the model’s planning capability and action stability.}
%补充关于chunk size大小的实验@cjy
Action chunking plays a pivotal role in manipulation tasks~\cite{act}. 
Training VLAs to predict action chunks implicitly endows them with planning capabilities and improves the temporal coherence of the generated actions. 
We employ this design and set the action chunking size to 5.
\begin{table}[H]
\caption{Comparison of different chunk sizes in terms of average length on CALVIN.}
\centering
\small % 或 \footnotesize 进一步缩小
\setlength{\tabcolsep}{2.3pt} % 减小列间距（默认是 6pt）

\label{tab: proprioception}
\begin{tabular}{ccccc}
\toprule
Chunk Size & 1 & 5 & 12 & 20 \\
\midrule
Avg. Len.  & 2.25 & \textbf{3.68} & 3.35 & 0.70 \\
\bottomrule
\end{tabular}
\end{table}

\subsection{Study on Training Curriculum (Q2)}
\label{subsec: training curriculum}
\noindent
% \textbf{Key Findings 4: 跨实例、跨环境的大规模预训练不是必须的，在域内多任务数据集上进行后训练即可建立视觉语言到动作的映射.}
%补充在oxe上预训练过的模型在robotwin上的实验，不用做全部任务，选三四个任务代表即可@暂时不做
%真机post training@sxq(已完成)
\textbf{Key Findings 5: Cross-embodiment large-scale pre-training is not essential.
Post-training on in-domain multi-task data is sufficient to establish the mapping from vision and language to action.}
%我们认为，跨构型的大规模公开数据集存在质量低、动作空间差异的问题，这些制约了在其上进行预训练的有效性。相对的，下游任务数据集往往由仿真中的高质量示范或人类专家采集的真实数据构成。因此，在下游数据集的多个任务上进行预训练，并在单个任务上进行微调是一个promising的做法。在双臂构型的真机和仿真实验中，我们在所有任务数据上进行预训练，然后在单个任务上进行微调，得到了最佳的性能。
Large-scale cross-embodiment pre-training often suffers from low-quality samples and discrepancies in action spaces, which limit its effectiveness. 
In contrast, domain-specific datasets are typically composed of high-quality demonstrations collected either in simulation or from human experts in real-world settings. 
Consequently, conducting post-training across multiple tasks within downstream datasets, followed by fine-tuning on a single task, emerges as a promising approach. 
In both real-world and simulated experiments with dual-arm configurations, we conduct the \textbf{two-stage} training paradigm, including post-training and fine-tuning, and make \method~achieves the best overall performance.
This suggests that the vision-to-action mapping can be effectively learned from domain-specific data when training tasks provide sufficient diversity in goals and scenes.
% We find that post-training on in-domain multitask datasets without cross-domain pre-training enables strong performance on real-world robotic tasks. This suggests that the vision-to-action mapping can be effectively learned from domain-specific data when training tasks provide sufficient diversity in goals and scenes. The results indicate that domain alignment and task coverage are more important than large-scale pre-training across environments.

\begin{table}[H]
\caption{Comparison of different training curricula on the seen tasks on RoboTwin.}
\centering
\small % 或 \footnotesize 进一步缩小
\setlength{\tabcolsep}{2.3pt} % 减小列间距（默认是 6pt）
\label{tab: training}
\begin{tabular}{cccccccc}
\toprule
% VLM & Param. &
% \multicolumn{5}{c}{Success Rate (\%)} & Avg. Len. \\
% method & projector & 1/5 & 2/5 & 3/5 & 4/5 & 5/5 & ABC$\rightarrow$D \\
% Training  &  \multicolumn{2}{c}{Success Rate (\%)} \\
Training Curriculum & Open Laptop & Lift Pot \\
\midrule
Pre-training & 20.0\% & 18.0\%  \\
Post-training & \textbf{38.0\%} & \textbf{39.0\%}  \\ 
\bottomrule
\end{tabular}
\end{table}

\begin{table*}[h]
\centering
\small
\caption{Comparison with various manipulation baselines on CALVIN.} 
\label{tab: calvin}
\scalebox{0.98}{
\begin{tabular}{cccccccccc}
\toprule
\multirow{2}{*}{Category} & \multirow{2}{*}{Method} & \multirow{2}{*}{Params} & w/o &
\multicolumn{5}{c}{Success Rate (\%)} & Avg. Len. \\
& & & Pre-training &  1/5 & 2/5 & 3/5& 4/5& 5/5 & ABC$\rightarrow$D \\
\midrule
\multirow{3}{*}{Generative Methods} 
& 3D-VLA~\cite{zhen20243d}~\textcolor{gray}{{(\textit{ICML'24})}} & 1B & $\times$ & 44.7 & 16.3 & 8.1 & 1.6 & 0 & 0.70\\
& GR-1~\cite{gr1}~\textcolor{gray}{{(\textit{ICLR'24})}} & 195M & $\times$  & 85.4& 71.2& 59.6& 49.7& 40.1& 3.06 \\
& Vidman~\cite{wen2024vidman}~\textcolor{gray}{{(\textit{NIPS'24})}} & 1B & $\times$  & 91.5 & 76.4 & 68.2 & 59.2 & 46.7& 3.42\\

% \midrule
% \multirow{1}{*}{3D Prediction Methods} 
% & RoboUniview &3B & \checkmark & 94.2 & 84.2 & \textbf{73.4} & 62.2& 50.7 & 3.65\\

\midrule
\multirow{1}{*}{Diffusion Policy} 
& 3D Diffuser Actor~\cite{ke20253d}~\textcolor{gray}{{(\textit{CoRL'24})}} & 70M & \checkmark  & 93.8 & 80.3 & 66.2 & 53.3 & 41.2& 3.35\\

\midrule
\multirow{3}{*}{Large VLA Models} 
& RoboFlamingo~\cite{li2024roboflamingo}~\textcolor{gray}{{(\textit{ICLR'24})}} & 3B & \checkmark  & 82.4& 61.9& 46.6& 33.1& 23.5 & 2.47 \\
& OpenVLA~\cite{kim2024openvla}~\textcolor{gray}{{(\textit{CoRL'24})}} & 7B & $\times$ & 91.3 & 77.8 & 62.0 & 52.1 & 43.5& 3.27\\
& \textbf{\method~(Ours)} & 500M & \checkmark & \textbf{94.8} & \textbf{84.5} & \textbf{71.3} & \textbf{62.5} & \textbf{53.8} & \textbf{3.68} \\
\bottomrule
\end{tabular}
}
\vspace{-2mm}
\end{table*}

% \subsection{Study on Parameter Amounts (Q3)}
% \textbf{Key Findings 5: 参数量与模型性能不成正比，通过上面的techniques，小型VLA性能可与大型VLA媲美.}
%把0.5b和7b的实验放过来@sxq(已完成)

\begin{table*}[t]
\caption{\textbf{Evaluation on RoboTwin benchmark.} Success rates for 8 tasks on the Seen and DR settings. Best result in each row highlighted in \textbf{Bold}. 
% RDT achieves the highest average in both settings (48.9\% for seen settings, 11.4\% for DR settings) under large models, while our method (\method) shows strong performance in the small model, especially in hard tasks (28.6\%).
}
\centering
\label{tab:performance}
\begin{tabular}{l c c c c c c c c c}
\toprule
&\multicolumn{6}{c}{Small Model w/o Pre-training} & \multicolumn{2}{c}{Large Model} \\
\midrule
\multirow{2}{*}{Simulation Task} & 
%\multicolumn{2}{c}{ACT~\cite{act}} & 
%\multicolumn{2}{c}{DP~\cite{dp}} & 
%\multicolumn{2}{c}{\method~} &  % \method in small model  {\method~}
%\multicolumn{2}{c}{~RDT\cite{rdt}} \\  % RDT in large model

\multicolumn{2}{c}{ACT} & 
\multicolumn{2}{c}{DP} & 
\multicolumn{2}{c}{\method~ (Ours)} &  % \method in small model  {\method~}
\multicolumn{2}{c}{RDT} \\  % RDT in large model

 & Seen & DR & Seen & DR & Seen & DR & Seen & DR \\
\midrule
Click Bell & 4.0\% & 2.0\% & 54.0\% & 0.0\% & 
\cellcolor{gray!20}\textbf{81.0\%} & \cellcolor{gray!20}\textbf{72.0\%} & 
80.0\% & 9.0\% \\
Click Alarmclock & 11.0\% & 0.0\% & 61.0\% & 5.0\% & 
\cellcolor{gray!20}\textbf{73.0\%} & \cellcolor{gray!20}\textbf{65.0\%} & 
61.0\% & 12.0\% \\
Lift Pot & 7.0\% & 2.0\% & 31.0\% & 0.0\% & 
39.0\% & \cellcolor{gray!20}\textbf{21.0\%} & 
\cellcolor{gray!20}\textbf{72.0\%} & 9.0\% \\
Move Can Pot & 0.0\% & 0.0\% & \cellcolor{gray!20}\textbf{39.0\%} & 0.0\% & 
28.0\% & \cellcolor{gray!20}\textbf{16.0\%} & 
25.0\% & 12.0\% \\
Open Laptop & 31.0\% & 0.0\% & 49.0\% & 0.0\% & 
38.0\% & \cellcolor{gray!20}\textbf{31.0\%} & 
\cellcolor{gray!20}\textbf{59.0\%} & 31.0\% \\
Pick Dual Bottles & 4.0\% & 0.0\% & 22.0\% & 0.0\% & 
26.0\% & 7.0\% & 
\cellcolor{gray!20}\textbf{41.0\%} & \cellcolor{gray!20}\textbf{10.0\%} \\
Place Dual Shoes & 0.0\% & 0.0\% & 7.0\% & 0.0\% & 
\cellcolor{gray!20}\textbf{8.0\%} & \cellcolor{gray!20}\textbf{5.0\%} & 
4.0\% & 3.0\% \\
Rotate Qrcode & 0.0\% & 0.0\% & 13.0\% & 0.0\% & 
29.0\% & \cellcolor{gray!20}\textbf{12.0\%} & 
\cellcolor{gray!20}\textbf{49.0\%} & 5.0\% \\
\midrule
\textbf{Average success} & 7.1\% & 0.5\% & 34.5\% & 0.6\% & 
40.3\% & \cellcolor{gray!20}\textbf{28.6\%} & 
\cellcolor{gray!20}\textbf{48.9\%} & 11.4\% \\
\bottomrule
\end{tabular}

\end{table*}

\subsection{Study on Action Space (Q3)}
%为了设计unified action space，我们首先研究其应该是离散的还是连续的，然后研究将navigation和manipulation统一的方式。
To design a unified action space, we first investigate whether it should be discrete or continuous (F6-7), and then explore how navigation and manipulation can be integrated within the same framework (F8).

\noindent
\textbf{Key Findings 6: Continuous action space in the diffusion head is not indispensable. 
With action chunking, discrete actions can achieve comparable performance.}
%different version of \method~
% \textbf{Key Findings 4: Limited Improvement with Diffusion Head for Continuous Action Generation} 
% while许多vla已经通过结合diffusion head to generate precise continuous actions，这种设计破坏了模型自回归的特性，从而限制了其与先进的vlm技术的结合的扩展性。
%我们通过action tokenization结合action chunking的方案，在继承自回归特性的前提下达到了有竞争力的性能。
While many VLAs adopt a diffusion head to generate precise continuous actions, this design compromises the autoregressive nature of the model, thereby limiting its scalability when integrated with advanced techniques on VLMs and LLMs. 
In contrast, our approach combines action tokenization with action chunking, achieving competitive performance while preserving the autoregressive property.

% The use of a diffusion head to generate continuous actions offers limited benefits and does not show significant improvement when compared to generating discrete actions using pure autoregressive methods. 
% Additionally, while the introduction of the diffusion head enhances the model's capability to handle continuous action spaces, it also increases the model's complexity. 
% This added complexity results in the loss of some causal reasoning abilities, as the model struggles to maintain the same level of efficient sequential decision-making inherent in autoregressive methods.

\begin{table}[H]
\centering
\small
\caption{Comparison among different action spaces of \method~on CALVIN.}
% , all models are built upon LLAMA-2-7b as the language backbone and CLIP as the vision encoder.}
\label{tab:actionhead}
\scalebox{0.98}{
\begin{tabular}{cccccccc}
\toprule
Decoder &
\multicolumn{5}{c}{Success Rate (\%)} & Avg. Len. \\
Type & 1/5 & 2/5 & 3/5 & 4/5 & 5/5 & ABC$\rightarrow$D \\
\midrule
Discrete & 94.8 & 84.5 & 71.3 & 62.5 & 53.8 & 3.68 \\
Continuous & 93.5 & 84.4 & 73.5 & 63.3 & 54.3 & 3.70 \\
\bottomrule
\end{tabular}
}
\end{table}

\noindent
\textbf{Key Findings 7: Fine-grained action tokenization does not lead to high performance.} 
% While a finer granularity in action representation might intuitively seem to improve the model's ability to capture subtle differences, it introduces more training complexity, which results in a loss of efficiency and generalization. 
While a finer granularity in action representation might intuitively seem to improve the model's ability to capture subtle differences, it introduces more training complexity to fit a larger action space. As shown in \Cref{tab: bin}, using a larger number of action bins leads to lower success rates compared with coarser discretization, indicating that overly fine-grained action tokenization results in a loss of efficiency and generalization. 
% The increased number of bins causes the model to overfit to the finer details of the action space, leading to poor performance on tasks that require broader action distinctions. Thus, using too many bins for action representation can hinder the model's ability to make effective predictions, as the model struggles to manage the additional complexity.
\begin{table}[H]
\caption{Comparison of different numbers of bins on CALVIN.}
\centering
\small % 或 \footnotesize 进一步缩小
\setlength{\tabcolsep}{2.3pt} % 减小列间距（默认是 6pt）
\label{tab: bin}
\begin{tabular}{cccccccc}
\toprule
% VLM & Param. &
% \multicolumn{5}{c}{Success Rate (\%)} & Avg. Len. \\
% method & projector & 1/5 & 2/5 & 3/5 & 4/5 & 5/5 & ABC$\rightarrow$D \\
Bin & \multicolumn{5}{c}{Success Rate (\%)} & Avg. Len. 
\\  Numbers  & 1/5 & 2/5 & 3/5 & 4/5 & 5/5 & ABC$\rightarrow$D \\
\midrule
256 & \textbf{94.8} & \textbf{84.5} & \textbf{71.3} & \textbf{62.5} & \textbf{53.8} & \textbf{3.68} \\
1024 & 96.4 & 86.5 & 72.1 & 60.2 & 48.6 & 3.65 \\ 
\bottomrule
\end{tabular}

\end{table}

% \subsection{Study on Hybrid Action Space (Q5)}
% 对于mobile manipulation任务，我们希望vla同时输出navagation数据和manipulation数据。一种直观的想法是将navigation数据与manipulation数据相同的方式进行tokenization，然而，我们在实验中发现这种方式不稳定，其往往容易导致机器人在停稳后的操纵过程中突然重新移动。Inspired by NaVid, 我们将导航部分的输出设置为一个direction token (前进、左转、右转、停止) 加一个衡量前进距离或转动角度的value token。特别的，为了保证manipulation过程中的稳定性，当direction为stop时，我们在其后输出manipulation的value token。这一设计允许模型在navigation和manipulation之间自由切换。
\noindent
\textbf{Key Findings 8: The unified action space can be realized through a combination of direction token and value token.}
For mobile manipulation tasks, we aim for the VLA to simultaneously output both navigation data and manipulation data. An intuitive approach is to tokenize the navigation data in the same manner as the manipulation data. 
However, experiments revealed that this method is unstable and often causes the robot to abruptly resume movement during the manipulation phase after coming to a stop. 
We designed the navigation output as a direction token (\textit{forward, turn left, turn right, stop}) followed by a value token representing the distance to advance or the angle to rotate. 
Specifically, to ensure stability during manipulation, when the direction token is \textit{stop}, we append the value tokens for manipulation after it. This design allows the model to flexibly switch between navigation and manipulation.
% Furthermore, we add a <Navigation> special token to 提示模型进行移动操纵任务。
Furthermore, we introduce a special \verb|<Navigation>| token following task instructions to prompt the model to perform mobile manipulation tasks.

\begin{table}[H]
\caption{Comparison of unified action space on the mobile manipulation tasks in the real world. 
% We report success rates (\%).
}
\centering
\small % 或 \footnotesize 进一步缩小
\setlength{\tabcolsep}{2.3pt} % 减小列间距（默认是 6pt）
\label{tab: training}
\begin{tabular}{cccccccc}
\toprule
% VLM & Param. &
% \multicolumn{5}{c}{Success Rate (\%)} & Avg. Len. \\
% method & projector & 1/5 & 2/5 & 3/5 & 4/5 & 5/5 & ABC$\rightarrow$D \\
Unified  &  Move and & Move and\\
Action Space &  fetch bottles &  open the drawer \\
\midrule
Action Value Token & 2/10 & 1/10  \\
Direction + Vaule Token & 4/10 & 4/10  \\ 
\bottomrule
\end{tabular}
\end{table}

\section{Evaluation of \method}
\label{sec:final}
 We comprehensively evaluate our final architecture on CEBench to evaluate its manipulation performance, capabilities of visual generalization, cross-embodiment versatility, and abilities of mobile manipulation. 
% \subsection{Experiment Setup}

% 所以实验都在8张H100上进行
\subsection{Training Setup}
All post-training is conducted on 8 NVIDIA H100 GPUs unless otherwise noted, and fine-tuning is conducted on 1 NVIDIA 4090 GPU.
For the CALVIN ABC$\rightarrow$D task split, we post-train on multiple tasks for a single epoch without fine-tuning, which costs approximately six hours. 
For bimanual manipulation on RoboTwin and in the real world, we perform 2 epochs of post-training followed by 8 epochs of fine-tuning. 
% For real-world bimanual manipulation, we pre-train for 2 epochs on a dataset of 2{,}000 trajectories, and then fine-tune for 8 epochs on a task-specific subset of 200 demonstrations.
\begin{table*}[t]  
\caption{Comparison of success rates on real-world bimanual tasks.}
\centering
\small
\setlength{\tabcolsep}{2pt}  % 紧凑列间距

\label{tab:real_results}
\begin{tabular}{l 
    c c | c c | c c | c c | c c | c c | c c} 
\toprule
Embodiments & \multicolumn{8}{c}{Basic Single-arm Tasks} & \multicolumn{4}{c}{Dexterous Bimanual Tasks}&& \\
\midrule
Task
& \multicolumn{2}{c}{Stack Bowls}
& \multicolumn{2}{c}{Restore Bottle}   
& \multicolumn{2}{c}{Click Bell} 
& \multicolumn{2}{c}{Place Vagetable}
& \multicolumn{2}{c}{Pack Bottles}     
& \multicolumn{2}{c}{Lift Pot} 
& Avg. &Success\\
Settings & Seen & DR & Seen & DR & Seen & DR & Seen & DR & Seen & DR  & Seen & DR & Seen &DR\\
\midrule
ACT~\cite{act} 
& 30\% & 10\% 
& 15\% & 0\%       
& 30\% & 10\% 
& 20\% & 0\% 
& 10\% & 0\%       
& 7\% & 0\% 
& 18.6\% &3.0\% \\

TinyVLA~\cite{tinyvla} 
& 25\% & 10\%   % Stack Bowls (Seen, DR)
& 10\% & 0\%    % Restore Bottle
& 25\% & 5\%    % Click Bell
& 15\% & 0\%    % Place Vagetable
& 20\% & 5\%    % Pack Bottles
& 10\% & 5\%    % Lift Pot
& 17.5\% & 4.2\%\\

\textbf{\method~(Ours)} 
& \textbf{58\%} & \textbf{40\%}   % +20
& \textbf{38\%} & \textbf{27\%}   % +20
& \textbf{66\%} & \textbf{54\%}   % +20
& \textbf{50\%} & \textbf{32\%}   % +20
& \textbf{28\%} & \textbf{16\%}   % +10
& \textbf{25\%} & \textbf{15\%}   % +10
& \textbf{44.2\%}  & \textbf{30.7\%}\\
\bottomrule
\end{tabular}
\label{tab: real_static}
\end{table*}

\begin{figure*}[t]
  \centering
  \includegraphics[width=0.99\textwidth]{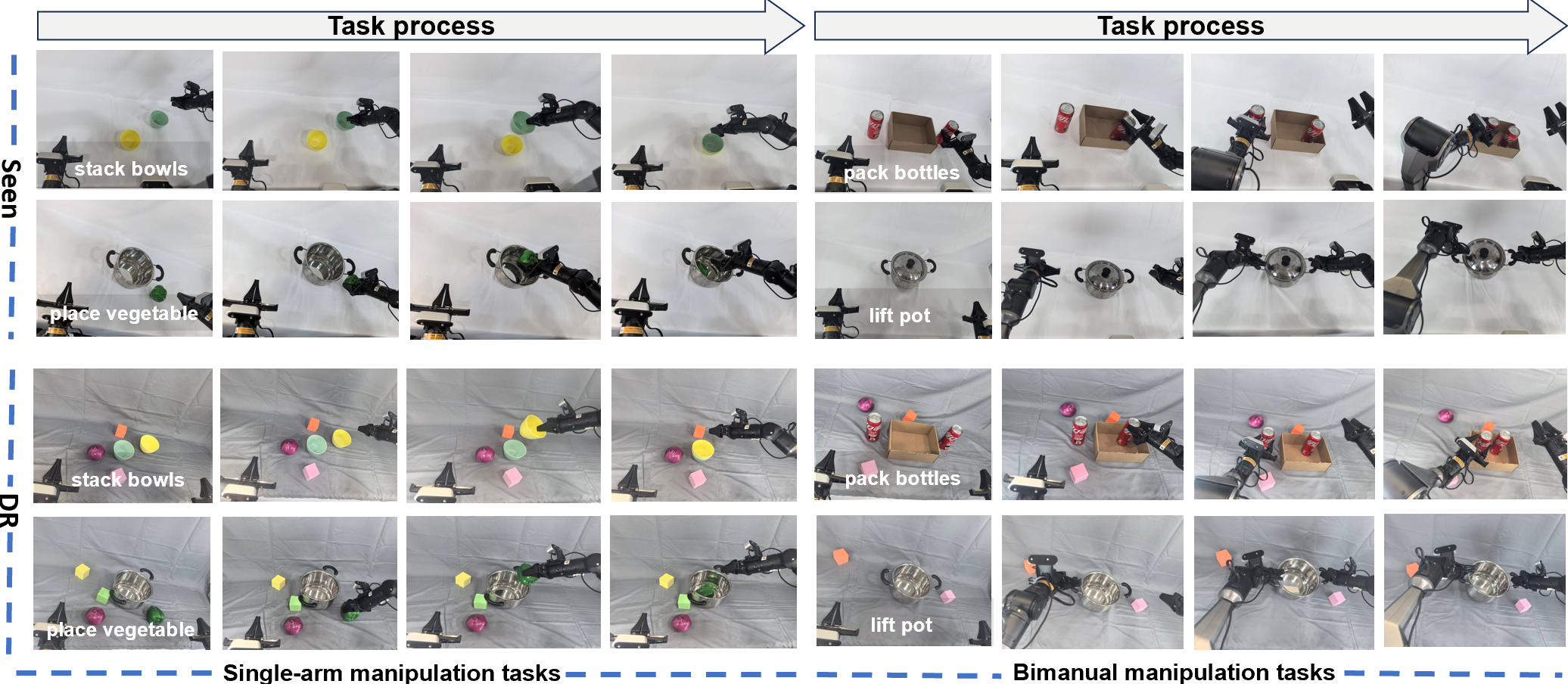}
  \caption{Visualization of real-world tasks.
  %上面两行展示了seen tasks，下面两行展示了settings with domain randomization. 我们从全部的8个真实世界任务重选择了四个作为例子。
  The top two rows illustrate the seen tasks, while the bottom two rows correspond to settings with domain randomization. 
  Out of the eight real-world tasks, we select two representative examples of single-arm manipulation (left) as well as two examples of bimanual collaboration (right).
  }
  \label{fig:wide-figure}
\end{figure*}

\subsection{Single-arm Manipulation on CALVIN}

\noindent\textbf{Evaluation detail.}
We report the average completed trajectory length (Avg. Len.) across all five subtasks as well as success rates on each subtask.
Following the official ABC$\rightarrow$D settings~\cite{calvin}, the evaluation is conducted in an unseen scene. 
% For a comprehensive comparison, we evaluate our approach against representative baselines~\cite{zhen20243d,gr1,wen2024vidman,ke20253d,li2024roboflamingo,kim2024openvla} that span multiple research domains. 
To ensure reliable evaluation, we test each method 1000 times.
% with a maximum of 360 steps per task.

% \begin{itemize}
%     \item Generative methods, including 3D-VLA~\cite{zhen20243d}, GR-1~\cite{gr1}, and Vidman\cite{wen2024vidman}, which explore sequence generation from language and visual inputs.
%     \item 3D prediction methods, exemplified by RoboUniview~\cite{liu2024robouniview}, which leverage structured 3D representations for action forecasting.
%     \item Diffusion-based policies, such as 3D Diffuser Actor~\cite{ke20253d}, which employ denoising diffusion techniques for visuomotor control.
%     \item Large-scale VLA models, including RoboFlamingo~\cite{li2024roboflamingo} and OpenVLA~\cite{kim2024openvla}, which represent recent efforts to scale multimodal pre-training for robotic manipulation.
% \end{itemize}

\noindent\textbf{Evaluation results.}
% As shown in \Cref{tab: calvin},~\method, built on a lightweight 0.5B LLM backbone without pre-training on large robot datasets, achieves the highest success rates in all sub-tasks, with an average completed length of 3.68, surpassing all compared baselines. 
\Cref{tab: calvin} shows that our~\method, with a lightweight 0.5B LLM and no large robot-dataset pre-training, tops all sub-tasks in success rate and attains the best average completed length of 3.68 against other Large VLA Models\cite{kim2024openvla,li2024roboflamingo}.
% Compared with strong methods such as RoboUniview and 3D Diffuser Actor,~\method~yields clear gains on the hardest subtasks (4/5 and 5/5). 
Compared with 3D-aware baselines\cite{ke20253d}, our~\method~outperforms them with a 0.33 increase in the average length of the completed trajectory, demonstrating its strong spatial reasoning ability.
Despite no pre-training, our model outperforms methods~\cite{wen2024vidman,gr1,zhen20243d} that rely on large-scale video pre-training and future image prediction.
These results show that combining our techniques enables a light model to outperform architecturally complex, large-parameter, and training-expensive models.

\subsection{Bimanual Manipulation on RoboTwin}

\noindent\textbf{Evaluation details.}
We evaluate our~\method~100 times in both the seen and DR settings and report success rates per task. 
For additional details, please refer to official settings~\cite{robotwin}.

\noindent\textbf{Evaluation results.}
% With quantitative results presented in TABLE I, \method~achieves a success rate of 40.3\% in the Easy setting and 28.6\% in the Hard setting. Although its performance in the Easy setting is slightly lower than that of RDT, in the Hard setting it makes a significant absolute improvement of +17.2\% over RDT~\cite{rdt}, +28.1\% over ACT~\cite{act}, and +28.0\% over DP~\cite{dp}, fully demonstrating its robustness and advantages in visually distracting environments.
% 从表中可以看出 在Small Model w/o Pre-training的设定下，我们的模型优于act dp等小模型，我们的架构优于融合了扩展或者vae的架构，对比参数量更大以及需要大规模预训练的模型，我们在tobotwin benchmark上的成功率也优于他们，证明了我们架构的有效性，可以弥补参数两以及训练样本带来的差距
\Cref{tab:performance}~shows that as a small model without pre-training, our \method~achieves a success rate of 40.3\% on seen tasks and 28.6\% on domain-randomization tasks, outperforming diffusion-based~\cite{dp} and VAE-based~\cite{act} baselines.
% , including ACT\cite{act} and DP~\cite{dp}.
Compared to larger VLAs that require extensive pretraining, our~\method~achieves higher success rates in the DR environment, which indicates that our \method~owns visual generalization as well as spatial comprehension capabilities.
% , although it uses far fewer parameters and no large-scale pretraining. 
% These results validate the effectiveness of our architecture and indicate that it can mitigate disadvantages in model size and training data scale, particularly on more challenging coordinated bimanual manipulation tasks.

\subsection{Bimanual Manipulation in the Real World}

\noindent\textbf{Evaluation details.}
For fixed-base bimanual manipulation, each method is evaluated over 100 episodes per task under both easy and hard settings, and for mobile manipulation, we report results over 10 episodes per task.

\noindent\textbf{Evaluation results on fixed-base tasks.}
\Cref{tab: real_static} shows that our~\method~consistently outperforms the ACT~\cite{act} and TinyVLA~\cite{tinyvla} across 6 tasks, demonstrating its effectiveness in real-world experiments.
%特别的，当面对背景变化和干扰物等unseen场景时，ACT和TinyVLA成功率大幅下降，接近0.而我们的方法受到的影响较小，这证明了我们的方法的视觉泛化能力。
Notably, when evaluated in unseen scenarios with background variations and distractor objects, both ACT and TinyVLA experience a dramatic drop in success rate, approaching zero. 
In contrast, \Cref{fig:wide-figure} shows that our~\method~is much less affected, demonstrating strong robustness and visual generalization capabilities.

\begin{figure}[t]  
    \centering
    \includegraphics[width=0.99\linewidth]{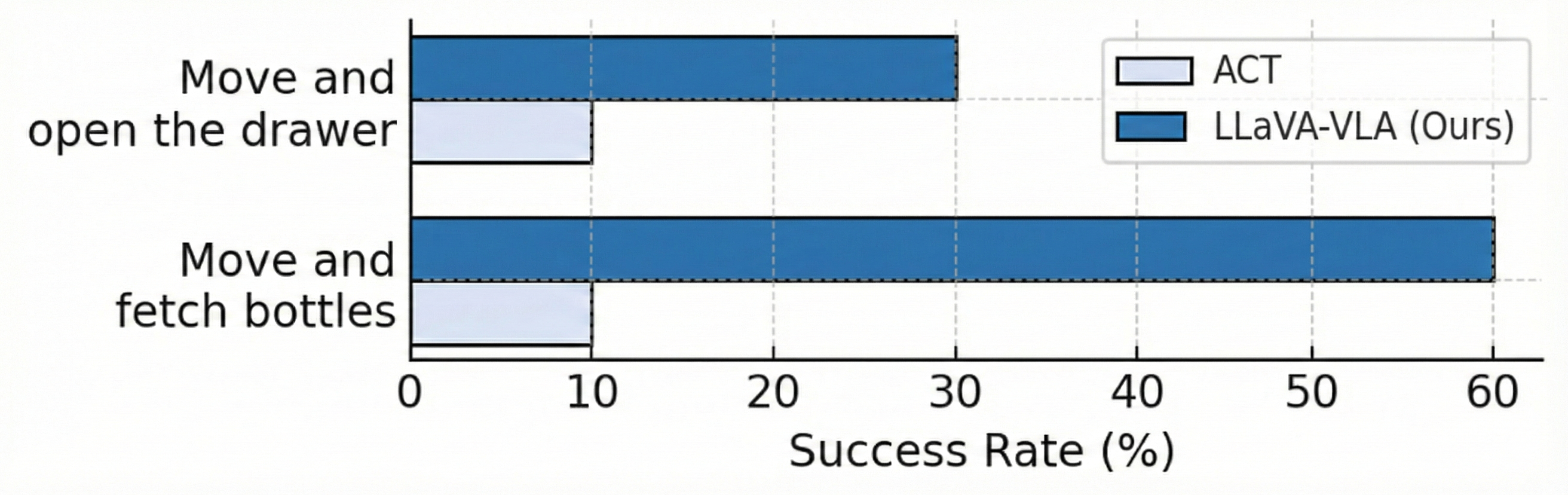} 
    \caption{Evaluation in real-world mobile manipulation tasks.}
    \label{fig:mobile_task} 
\end{figure}

\noindent\textbf{Evaluation results on mobile manipulation tasks.}
%由于existing VLAs都不具备移动操纵能力，因此我们选择ACT在mobile ALOHA中的实现作为baseline。实验证明ACT的navigation精度较低，进而无法正常进行操纵任务，仅有10%的成功率。同时，其不具备多任务学习的能力。而我们的方法利用SLM的能力，从多任务的广泛轨迹上进行学习，从而实现了指令跟随和精准的mobile manipulation。
Since existing VLA models lack mobile manipulation capabilities, we adopt the implementation of ACT in the mobile ALOHA~\cite{mobilealoha} setting as the baseline. However, \Cref{fig:mobile_task} shows that ACT suffers from low navigation accuracy, which prevents it from reliably executing manipulation tasks, resulting in only a 10\% success rate. 
Moreover, ACT is not equipped with multi-task learning capabilities. 
In contrast, our approach leverages the strengths of small VLMs to learn from diverse multi-task trajectories, thereby enabling effective instruction following and precise mobile manipulation.

\section{Conclusion}
In this work, we introduced CEBench, a practical benchmark that spans diverse embodiments across both simulation and the real world and explicitly considers potential domain randomization. 
By conducting extensive experiments on CEBench, we systematically investigated the design space of lightweight VLAs and distilled several key insights into their architecture, training, and action representation. 
Building upon these insights, we developed~\method, a lightweight yet powerful VLA architecture that eliminates the need for costly pre-training while maintaining strong performance. 
% Building upon these insights, we developed~\method, a lightweight yet powerful VLA architecture that removes costly pre-training while maintaining strong performance. 
The experimental results demonstrate its cross-embodiment versatility and visual generalization. 
Meanwhile, the real-world evaluations establish~\method~as the first end-to-end VLA model capable of mobile manipulation. 
Our study builds a road map by providing valuable insights toward making VLA research more practical and accessible for embodied robotic systems. 
% We will release all datasets, codes, and checkpoints to accelerate progress in this emerging field.  
% \section{Acknowledgments}
% This work was supported by the Brain Science and Brain-like Intelligence Technology — National Science and Technology Major Project (Grant No. 2022ZD0208800)
% \newpage
\bibliographystyle{IEEEtran}
\bibliography{root}

@article{PD-VLA,
  title={Accelerating Vision-Language-Action Model Integrated with Action Chunking via Parallel Decoding},
  author={Song, Wenxuan and Chen, Jiayi and Ding, Pengxiang and Zhao, Han and Zhao, Wei and Zhong, Zhide and Ge, Zongyuan and Ma, Jun and Li, Haoang},
  journal={arXiv preprint arXiv:2503.02310},
  year={2025}
}

@article{CEED-VLA,
  title={CEED-VLA: Consistency Vision-Language-Action Model with Early-Exit Decoding},
  author={Song, Wenxuan and Chen, Jiayi and Ding, Pengxiang and Huang, Yuxin and Zhao, Han and Wang, Donglin and Li, Haoang},
  journal={arXiv preprint arXiv:2506.13725},
  year={2025}
}

@inproceedings{prismatic-vlms,
  title={Prismatic vlms: Investigating the design space of visually-conditioned language models},
  author={Karamcheti, Siddharth and Nair, Suraj and Balakrishna, Ashwin and Liang, Percy and Kollar, Thomas and Sadigh, Dorsa},
  booktitle={Forty-first International Conference on Machine Learning},
  year={2024}
}

@inproceedings{kim2024openvla,
  title={OpenVLA: An Open-Source Vision-Language-Action Model},
  author={Kim, Moo Jin and Pertsch, Karl and Karamcheti, Siddharth and Xiao, Ted and Balakrishna, Ashwin and Nair, Suraj and Rafailov, Rafael and Foster, Ethan P and Sanketi, Pannag R and Vuong, Quan and others},
  booktitle={8th Annual Conference on Robot Learning}
}

@inproceedings{quarvla,
  title={Quar-vla: Vision-language-action model for quadruped robots},
  author={Ding, Pengxiang and Zhao, Han and Zhang, Wenjie and Song, Wenxuan and Zhang, Min and Huang, Siteng and Yang, Ningxi and Wang, Donglin},
  booktitle={European Conference on Computer Vision},
  pages={352--367},
  year={2024},
  organization={Springer}
}

@inproceedings{li2024roboflamingo,
  title={Vision-Language Foundation Models as Effective Robot Imitators},
  author={Li, Xinghang and Liu, Minghuan and Zhang, Hanbo and Yu, Cunjun and Xu, Jie and Wu, Hongtao and Cheang, Chilam and Jing, Ya and Zhang, Weinan and Liu, Huaping and others},
  booktitle={The Twelfth International Conference on Learning Representations}
}

@misc{llama,
      title={LLaMA: Open and Efficient Foundation Language Models}, 
      author={Hugo Touvron and Thibaut Lavril and Gautier Izacard and others},
      year={2023},
      eprint={2302.13971},
      archivePrefix={arXiv},
      primaryClass={cs.CL},
      url={https://arxiv.org/abs/2302.13971}, 
}

@inproceedings{oft,
  title={Fine-Tuning Vision-Language-Action Models: Optimizing Speed and Success},
  author={Kim, Moo Jin and Finn, Chelsea and Liang, Percy},
  journal={Robotics: Science and Systems},
  year={2025}
}

@article{robovlm,
  title={Towards Generalist Robot Policies: What Matters in Building Vision-Language-Action Models},
  author={Li, Xinghang and Li, Peiyan and Liu, Minghuan and Wang, Dong and Liu, Jirong and Kang, Bingyi and Ma, Xiao and Kong, Tao and Zhang, Hanbo and Liu, Huaping},
  journal={arXiv preprint arXiv:2412.14058},
  year={2024}
}

@article{black2024pi_0,
  title={$\pi_0 $: A Vision-Language-Action Flow Model for General Robot Control},
  author={Black, Kevin and Brown, Noah and Driess, Danny and Esmail, Adnan and Equi, Michael and Finn, Chelsea and Fusai, Niccolo and Groom, Lachy and Hausman, Karol and Ichter, Brian and others},
  journal={arXiv preprint arXiv:2410.24164},
  year={2024}
}

@article{bai2023qwen,
  title={Qwen-vl: A frontier large vision-language model with versatile abilities},
  author={Bai, Jinze and Bai, Shuai and Yang, Shusheng and Wang, Shijie and Tan, Sinan and Wang, Peng and Lin, Junyang and Zhou, Chang and Zhou, Jingren},
  journal={arXiv preprint arXiv:2308.12966},
  year={2023}
}

@article{hu2021lora,
  title={Lora: Low-rank adaptation of large language models},
  author={Hu, Edward J and Shen, Yelong and Wallis, Phillip and Allen-Zhu, Zeyuan and Li, Yuanzhi and Wang, Shean and Wang, Lu and Chen, Weizhu},
  journal={International Conference on Learning Representations (ICLR)},
  year={2021}
}

@inproceedings{clip,
  title={Learning transferable visual models from natural language supervision},
  author={Radford, Alec and Kim, Jong Wook and Hallacy, Chris and Ramesh, Aditya and Goh, Gabriel and Agarwal, Sandhini and Sastry, Girish and Askell, Amanda and Mishkin, Pamela and Clark, Jack and others},
  booktitle={International conference on machine learning},
  pages={8748--8763},
  year={2021},
  organization={PmLR}
}

@article{gr1,
      title={Unleashing Large-Scale Video Generative Pre-training for Visual Robot Manipulation}, 
      author={Hongtao Wu and Ya Jing and Chilam Cheang and Guangzeng Chen and Jiafeng Xu and Xinghang Li and Minghuan Liu and Hang Li and Tao Kong},
      journal={ICLR},
      year={2024}
}

@article{calvin,
  title={CALVIN: A Benchmark for Language-Conditioned Policy Learning for Long-Horizon Robot Manipulation Tasks},
  author={Oier Mees and Luk{\'a}s Hermann and Erick Rosete-Beas and Wolfram Burgard},
  journal={IEEE Robotics and Automation Letters},
  year={2021}
}

@article{rt1,
  title={RT-1: Robotics Transformer for Real-World Control at Scale},
  author={Anthony Brohan and Noah Brown and Justice Carbajal and Yevgen Chebotar and Joseph Dabis and Chelsea Finn and and others},
  journal={Proceedings of Robotics: Science and Systems},
  year={2023},
}

@article{fast,
  title={Fast: Efficient action tokenization for vision-language-action models},
  author={Pertsch, Karl and Stachowicz, Kyle and Ichter, Brian and Driess, Danny and Nair, Suraj and Vuong, Quan and Mees, Oier and Finn, Chelsea and Levine, Sergey},
  journal={arXiv preprint arXiv:2501.09747},
  year={2025}
}

@article{robotwin,
  title={RoboTwin 2.0: A Scalable Data Generator and Benchmark with Strong Domain Randomization for Robust Bimanual Robotic Manipulation},
  author={Chen, Tianxing and Chen, Zanxin and Chen, Baijun and Cai, Zijian and Liu, Yibin and Liang, Qiwei and Li, Zixuan and Lin, Xianliang and Ge, Yiheng and Gu, Zhenyu and others},
  journal={arXiv preprint arXiv:2506.18088},
  year={2025}
}

@inproceedings{act,
  title={Learning fine-grained bimanual manipulation with low-cost hardware},
  author={Zhao, Tony Z and Kumar, Vikash and Levine, Sergey and Finn, Chelsea},
  journal={Robotics: Science and Systems (RSS)},
  year={2023}
}

@article{dp,
  title={Diffusion policy: Visuomotor policy learning via action diffusion},
  author={Chi, Cheng and Xu, Zhenjia and Feng, Siyuan and Cousineau, Eric and Du, Yilun and Burchfiel, Benjamin and Tedrake, Russ and Song, Shuran},
  journal={The International Journal of Robotics Research},
  pages={02783649241273668},
  year={2023},
  publisher={SAGE Publications Sage UK: London, England}
}

@inproceedings{rdt,
  title={RDT-1B: a Diffusion Foundation Model for Bimanual Manipulation},
  author={Liu, Songming and Wu, Lingxuan and Li, Bangguo and Tan, Hengkai and Chen, Huayu and Wang, Zhengyi and Xu, Ke and Su, Hang and Zhu, Jun},
  booktitle={The Thirteenth International Conference on Learning Representations},
  year={2024}
}

@article{minigpt-4,
  title={Minigpt-4: Enhancing vision-language understanding with advanced large language models},
  author={Zhu, Deyao and Chen, Jun and Shen, Xiaoqian and Li, Xiang and Elhoseiny, Mohamed},
  journal={12th International Conference on Learning Representations, ICLR 2024},
  year={2024}
}

@article{llava,
  title={Visual instruction tuning},
  author={Liu, Haotian and Li, Chunyuan and Wu, Qingyang and Lee, Yong Jae},
  journal={Advances in neural information processing systems},
  volume={36},
  pages={34892--34916},
  year={2023}
}

@article{gemini,
  title={Gemini: A family of highly capable multimodal models, 2024},
  author={Team, Gemini and Anil, R and Borgeaud, S and Wu, Y and Alayrac, JB and Yu, J and Soricut, R and Schalkwyk, J and Dai, AM and Hauth, A and others},
  journal={arXiv preprint arXiv:2312.11805},
  volume={10},
  year={2024}
}

@article{tinyvla,
  title={Tinyvla: Towards fast, data-efficient vision-language-action models for robotic manipulation},
  author={Wen, Junjie and Zhu, Yichen and Li, Jinming and Zhu, Minjie and Tang, Zhibin and Wu, Kun and Xu, Zhiyuan and Liu, Ning and Cheng, Ran and Shen, Chaomin and others},
  journal={IEEE Robotics and Automation Letters},
  year={2025},
  publisher={IEEE}
}

@article{smolvla,
  title={Smolvla: A vision-language-action model for affordable and efficient robotics},
  author={Shukor, Mustafa and Aubakirova, Dana and Capuano, Francesco and Kooijmans, Pepijn and Palma, Steven and Zouitine, Adil and Aractingi, Michel and Pascal, Caroline and Russi, Martino and Marafioti, Andres and others},
  journal={arXiv preprint arXiv:2506.01844},
  year={2025}
}

@inproceedings{xiang2020sapien,
  title={Sapien: A simulated part-based interactive environment},
  author={Xiang, Fanbo and Qin, Yuzhe and Mo, Kaichun and Xia, Yikuan and Zhu, Hao and Liu, Fangchen and Liu, Minghua and Jiang, Hanxiao and Yuan, Yifu and Wang, He and others},
  booktitle={Proceedings of the IEEE/CVF conference on computer vision and pattern recognition},
  pages={11097--11107},
  year={2020}
}

@article{hung2025nora,
  title={Nora: A small open-sourced generalist vision language action model for embodied tasks},
  author={Hung, Chia-Yu and Sun, Qi and Hong, Pengfei and Zadeh, Amir and Li, Chuan and Tan, U and Majumder, Navonil and Poria, Soujanya and others},
  journal={arXiv preprint arXiv:2504.19854},
  year={2025}
}

@inproceedings{zhen20243d,
  title={3D-VLA: a 3D vision-language-action generative world model},
  author={Zhen, Haoyu and Qiu, Xiaowen and Chen, Peihao and Yang, Jincheng and Yan, Xin and Du, Yilun and Hong, Yining and Gan, Chuang},
  booktitle={Proceedings of the 41st International Conference on Machine Learning},
  pages={61229--61245},
  year={2024}
}

@article{wen2024vidman,
  title={Vidman: Exploiting implicit dynamics from video diffusion model for effective robot manipulation},
  author={Wen, Youpeng and Lin, Junfan and Zhu, Yi and Han, Jianhua and Xu, Hang and Zhao, Shen and Liang, Xiaodan},
  journal={Advances in Neural Information Processing Systems},
  volume={37},
  pages={41051--41075},
  year={2024}
}

@article{llavaov,
  title={Llava-onevision: Easy visual task transfer},
  author={Li, Bo and Zhang, Yuanhan and Guo, Dong and Zhang, Renrui and Li, Feng and Zhang, Hao and Zhang, Kaichen and Zhang, Peiyuan and Li, Yanwei and Liu, Ziwei and others},
  journal={arXiv preprint arXiv:2408.03326},
  year={2024}
}

@inproceedings{ke20253d,
  title={3D Diffuser Actor: Policy Diffusion with 3D Scene Representations},
  author={Ke, Tsung-Wei and Gkanatsios, Nikolaos and Fragkiadaki, Katerina},
  booktitle={Conference on Robot Learning},
  pages={1949--1974},
  year={2025},
  organization={PMLR}
}

@misc{belkhale2024minivla,
  title={Minivla: A better vla with a smaller footprint},
  author={Belkhale, Suneel and Sadigh, Dorsa},
  year={2024},
  publisher={Stanford-ILIAD GitHub: openvla-mini}
}

@inproceedings{mobilealoha,
  author    = {Fu, Zipeng and Zhao, Tony Z. and Finn, Chelsea},
  title     = {Mobile ALOHA: Learning Bimanual Mobile Manipulation with Low-Cost Whole-Body Teleoperation},
  booktitle = {{Conference on Robot Learning (CoRL)}},
  year      = {2024},
}

@inproceedings{siglip,
  title={Sigmoid loss for language image pre-training},
  author={Zhai, Xiaohua and Mustafa, Basil and Kolesnikov, Alexander and Beyer, Lucas},
  booktitle={Proceedings of the IEEE/CVF international conference on computer vision},
  pages={11975--11986},
  year={2023}
}

@article{achiam2023gpt,
  title={Gpt-4 technical report},
  author={Achiam, Josh and Adler, Steven and Agarwal, Sandhini and Ahmad, Lama and Akkaya, Ilge and Aleman, Florencia Leoni and Almeida, Diogo and Altenschmidt, Janko and Altman, Sam and Anadkat, Shyamal and others},
  journal={arXiv preprint arXiv:2303.08774},
  year={2023}
}

@article{reconvla,
  title={ReconVLA: Reconstructive Vision-Language-Action Model as Effective Robot Perceiver},
  author={Song, Wenxuan and Zhou, Ziyang and Zhao, Han and Chen, Jiayi and Ding, Pengxiang and Yan, Haodong and Huang, Yuxin and Tang, Feilong and Wang, Donglin and Li, Haoang},
  journal={arXiv preprint arXiv:2508.10333},
  year={2025}
}

@article{openhelix,
  title={Openhelix: A short survey, empirical analysis, and open-source dual-system vla model for robotic manipulation},
  author={Cui, Can and Ding, Pengxiang and Song, Wenxuan and Bai, Shuanghao and Tong, Xinyang and Ge, Zirui and Suo, Runze and Zhou, Wanqi and Liu, Yang and Jia, Bofang and others},
  journal={arXiv preprint arXiv:2505.03912},
  year={2025}
}

@article{zhong2025flowvla,
  title={FlowVLA: Thinking in Motion with a Visual Chain of Thought},
  author={Zhong, Zhide and Yan, Haodong and Li, Junfeng and Liu, Xiangchen and Gong, Xin and Song, Wenxuan and Chen, Jiayi and Li, Haoang},
  journal={arXiv preprint arXiv:2508.18269},
  year={2025}
}

@article{song2025rationalvla,
  title={Rationalvla: A rational vision-language-action model with dual system},
  author={Song, Wenxuan and Chen, Jiayi and Li, Wenxue and He, Xu and Zhao, Han and Cui, Can and Su, Pengxiang Ding Shiyan and Tang, Feilong and Cheng, Xuelian and Wang, Donglin and others},
  journal={arXiv preprint arXiv:2506.10826},
  year={2025}
}

@article{bjorck2025gr00t,
  title={Gr00t n1: An open foundation model for generalist humanoid robots},
  author={Bjorck, Johan and Casta{\~n}eda, Fernando and Cherniadev, Nikita and Da, Xingye and Ding, Runyu and Fan, Linxi and Fang, Yu and Fox, Dieter and Hu, Fengyuan and Huang, Spencer and others},
  journal={arXiv preprint arXiv:2503.14734},
  year={2025}
}

@article{spatialforcing,  
title     = {Spatial Forcing: Implicit Spatial Representation Alignment For Vision-Language-Action Model},
  author    = {Li, Fuhao and Song, Wenxuan and Zhao, Han and Wang, Jingbo and Ding, Pengxiang and Wang, Donglin and Zeng, Long and Li, Haoang},
  journal   = {arXiv preprint arXiv:2510.12276},
  year      = {2025},
}

@article{udvla,
title={Unified Diffusion VLA: Vision-Language-Action Model via Joint Discrete Denoising Diffusion Process},
author={Jiayi Chen and Wenxuan Song and Pengxiang Ding and Ziyang Zhou and Han Zhao and Feilong Tang and Donglin Wang and Haoang Li},
year={2025},
journal={arXiv preprint arXiv:2511.01718}
}

@article{upvla,
  title={Up-vla: A unified understanding and prediction model for embodied agent},
  author={Zhang, Jianke and Guo, Yanjiang and Hu, Yucheng and Chen, Xiaoyu and Zhu, Xiang and Chen, Jianyu},
  journal={arXiv preprint arXiv:2501.18867},
  year={2025}
}

@article{vla-adapter,
  author={Wang, Yihao and Ding, Pengxiang and Li, Lingxiao and Cui, Can and Ge, Zirui and Tong, Xinyang and Song, Wenxuan and Zhao, Han and Zhao, Wei and Hou, Pengxu and Huang, Siteng and Tang, Yifan and Wang, Wenhui and Zhang, Ru and Liu, Jianyi and Wang, Donglin},
  title={VLA-Adapter: An Effective Paradigm for Tiny-Scale Vision-Language-Action Model},
  journal={arXiv preprint arXiv:2509.09372},
  year={2025}
}

@article{zhang2026vlm4vla,
  title={VLM4VLA: Revisiting Vision-Language-Models in Vision-Language-Action Models},
  author={Zhang, Jianke and Chen, Xiaoyu and Wang, Qiuyue and Li, Mingsheng and Guo, Yanjiang and Hu, Yucheng and Zhang, Jiajun and Bai, Shuai and Lin, Junyang and Chen, Jianyu},
  journal={arXiv preprint arXiv:2601.03309},
  year={2026}
}
% \begin{thebibliography}{99}

% \bibitem{c17} J. G. Kreifeldt, ÒAn analysis of surface-detected EMG as an amplitude-modulated noise,Ó presented at the 1989 Int. Conf. Medicine and Biological Engineering, Chicago, IL.
% \bibitem{c18} J. Williams, ÒNarrow-band analyzer (Thesis or Dissertation style),Ó Ph.D. dissertation, Dept. Elect. Eng., Harvard Univ., Cambridge, MA, 1993. 
% \bibitem{c19} N. Kawasaki, ÒParametric study of thermal and chemical nonequilibrium nozzle flow,Ó M.S. thesis, Dept. Electron. Eng., Osaka Univ., Osaka, Japan, 1993.
% \bibitem{c20} J. P. Wilkinson, ÒNonlinear resonant circuit devices (Patent style),Ó U.S. Patent 3 624 12, July 16, 1990. 

% \end{thebibliography}

\end{document}